\documentclass[conference]{IEEEtran}
\IEEEoverridecommandlockouts
\usepackage{cite}
\usepackage{amsmath,amssymb,amsfonts}
\usepackage{algorithmic}
\usepackage{graphicx}
\usepackage{textcomp}
\usepackage{wrapfig}
\usepackage{subcaption}
\usepackage{xcolor}
\usepackage{hyperref}
\def\BibTeX{{\rm B\kern-.05em{\sc i\kern-.025em b}\kern-.08em
    T\kern-.1667em\lower.7ex\hbox{E}\kern-.125emX}}
    
\begin{document}


\title{TexControl: Sketch-Based Two-Stage Fashion Image Generation Using Diffusion Model}

\author{\IEEEauthorblockN{Yongming Zhang, Tianyu Zhang and Haoran Xie\IEEEauthorrefmark{1}}
\IEEEauthorblockA{Japan Advanced Institute of Science and Technology \\
Ishikawa, Japan}}


\maketitle

\newcommand\blfootnote[1]{
  \begingroup
  \renewcommand\thefootnote{}\footnote{#1}%
  \addtocounter{footnote}{-1}%
  \endgroup
}
\blfootnote{\IEEEauthorrefmark{1}Corresponding author (xie@jaist.ac.jp).} 

\begin{abstract}
Deep learning-based sketch-to-clothing image generation provides the initial designs and inspiration in the fashion design processes. However, clothing generation from freehand drawing is challenging due to the sparse and ambiguous information from the drawn sketches. The current generation models may have difficulty generating detailed texture information. In this work, we propose TexControl, a sketch-based fashion generation framework that uses a two-stage pipeline to generate the fashion image corresponding to the sketch input. First, we adopt ControlNet to generate the fashion image from sketch and keep the image outline stable. Then, we use an image-to-image method to optimize the detailed textures of the generated images and obtain the final results. The evaluation results show that TexControl can generate fashion images with high-quality texture as fine-grained image generation.
\end{abstract}


\section{Introduction}

Fashion design holds practical significance in both culture and artistic expression. Deep learning-based clothing generation methods significantly contribute to the field of art creation, including sketch-to-clothing 3D model generation\cite{he2023sketch2cloth}, and clothing image generation\cite{jain2019text}. In recent advancements in diffusion models\cite{saharia2022photorealistic,ramesh2022hierarchical,rombach2022high}, novel text-to-image generation methods have showcased remarkable image quality and brought new methods to the fashion design task. However, clothing generation diffusion models currently face a key issue: diffusion models make it difficult to generate clothing images with high-quality texture and exact material. 


\begin{figure}[t]
    \centering
    \includegraphics[width=0.95\linewidth]{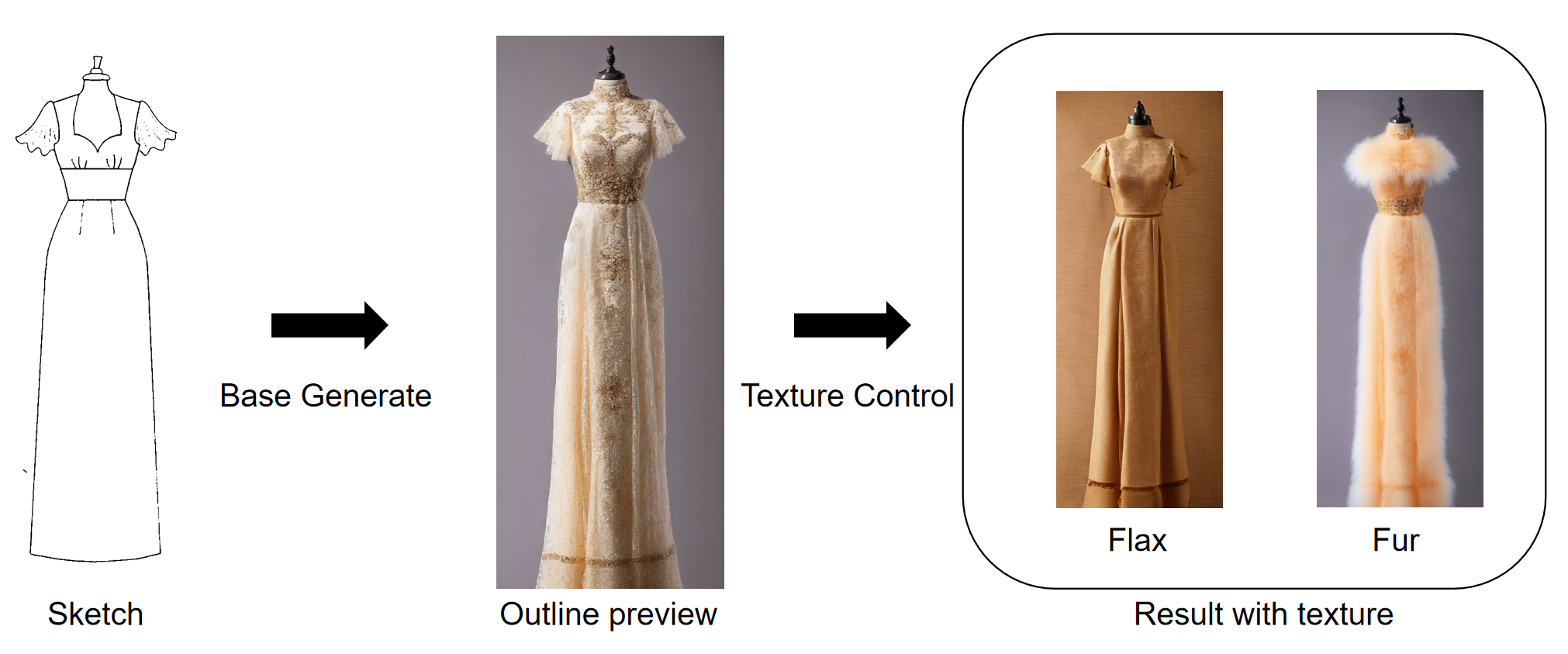}
    \caption{The proposed method, TexControl, adopts sketches as conditional input and generates fine-designed clothing images whose textures are consistent with the text inputs. The outline preview images are applied to divide TexControl into two stages: sketch-to-image stage and image-to-image stage.}
    \label{fig:overview}
\end{figure}

To solve this issue, We leverage a two-stage model to decompose the complex task of generating controllable clothing images into two simpler sub-tasks: outline consistency and texture control. Utilizing the two-stage model enables users to independently optimize the results of each stage, thereby providing high-quality outcomes. In addition, we use the sketch as conditioning input to provide the outline guidance for clothing image generation. Sketch enables the intuitive and concise expression of target object details, such as clothing patterns and accessories. 


\begin{figure*}[ht]
    \centering
    \includegraphics[width=0.95\linewidth]{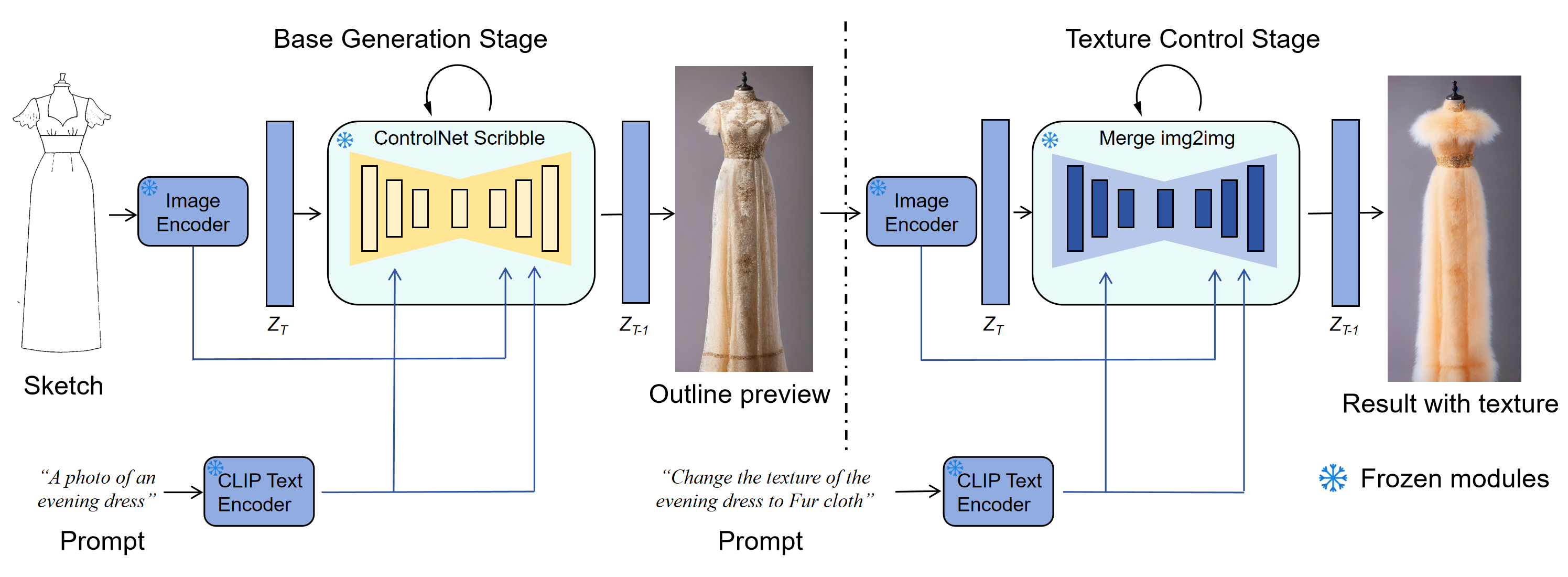}
    \caption{The framework of TexControl. TexControl consists of two stages: The base generation stage uses the ControlNet Scribble to generate an outline preview, and the texture control stage uses the ControlNet ip2p with model merge to generate the fine-designed result. $Z_T$ is the latent representation in latent space while $T$ is the timesteps.}
    \label{fig:framework}
\end{figure*}

In this work, we propose TexControl to generate clothing images with the required texture from hand-drawn sketches. As illustrated in Fig.~\ref{fig:overview}, the input of TexControl is hand-drawn sketches and text, while the output is fine-designed clothing images. We use the outline previews to divide the TexControl into two stages. In the base generation stage, sketch-to-image ControlNet\cite{zhang2023adding} is applied to generate outline previews from freehand sketches, which can accurately capture the outline of the input sketch and retain the features in the generated results to the greatest extent. In the texture control stage, TexControl uses the ControlNet image-to-image model to generate fine-designed results from the outline previews, whose texture corresponds with the text input. To verify our method, we conducted a comparison study between previous sketch-to-image generation methods and the TexControl. The evaluation results demonstrated the superiority of TexControl in generating comprehensive and detailed clothing images.  



\section{Related Work}

\subsection{Diffusion Models}

The diffusion model has demonstrated image quality surpassing that of GANs and VAEs, and its development has accelerated significantly in recent years. Which was initially introduced by Sohl-Dickstein et al.\cite{sohl2015deep} and later advanced by Song et al.\cite{song2019generative} and Ho et al.\cite{ho2020denoising}. 

The diffusion model primarily consists of two processes: the diffusion process and the denoising process. In the diffusion process, the diffusion model destroys the initial samples by continuously introducing Gaussian noise. In the denoising process, the diffusion model reconstructs the initial samples from severely disturbed data, thereby learning this denoising process. As the diffusion model was developed, researchers tried to optimize the diffusion model in two directions: image quality and generation speed. Among the famous models are the Latent Diffusion Model (LDM)\cite{rombach2022high} and the Latent Consistency Model (LCM)\cite{luo2023latent}. LDM reduces the dimensionality of the data by projecting it into a low-dimensional latent space, thereby reducing the computational complexity. LCM can directly map any step on the time schedule to the initial sample through a function, thus eliminating the complex iterative process and enhancing the speed of image generation.

\subsection{Clothing Image Generation}

The objective of clothing image generation is to visually demonstrate the design impact of garments under specified conditions or inputs (such as sketches, prompts, or style references), eliminating the need for physical samples. Previous works have generated high-quality clothing images by simulating the texture, material, and shape of the fabric, thus playing a significant role in the field of fashion design. 

FashionGAN\cite{cui2018fashiongan} introduced an end-to-end clothing image generation method based on cGAN, that quickly and automatically generates images from sketches and specified fabric images. Text2Human\cite{jiang2022text2human} presented a two-stage method to synthesize full-body human images from given poses and texts. Particularly, Text2Human generates clothing textures of high quality with fine-grained textual input. Recently, diffusion models have greatly improved the quality of generated images. Multimodal Garment Designer\cite{baldrati2023multimodal} proposed a multimodal fashion image editing method based on the latent diffusion model, allowing users to generate the garment images following multimodal prompts. In general, the diffusion model delivers high-quality results with challenges in controllability. Nevertheless, it represents a novel and promising method in the fashion domain.

\section{Proposed Method}


We discuss the detailed composition of our proposed two-stage sketch-to-clothing image generation method in this section. We first give a preliminary on ControlNet in Section~\ref{sec:controlnet}. We then introduce the two stages in our proposed model (as shown in Fig.~\ref{fig:framework}). In the base generation stage,  sketch-to-image ControlNet is utilized to generate outline previews from input sketches (introduced in Section~\ref{sec:1st_stage}). In the texture control stage,  image-to-image ControlNet is applied to generate the fine-designed results from middle products (introduced in Section~\ref{sec:2nd_stage}).

\subsection{Preliminary on ControlNet}
\label{sec:controlnet}

ControlNet\cite{zhang2023adding}, as an improvement upon the base LDM, allows for the incorporation of specific conditional inputs through fine-tuning into the pre-trained text-to-image diffusion models.

Since LDM is essentially a U-Net with an encoder, a middle block, and a skip-connected decoder. Suppose $\mathcal{F} \left( \cdot ;\Theta \right)$ is pre-trained LDM with parameters $\Theta$, the output feature maps $\boldsymbol{y}$ will be transformed by input feature maps $\boldsymbol{x}$:

\begin{equation}
    \boldsymbol{y} = \mathcal{F} \left( \boldsymbol{x};\Theta \right)
    \label{equ:1}
\end{equation}

ControlNet freezes the initial weights $\Theta$ in the LDM and builds a copy of the encoder and middle blocks, called trainable copy. The parameters $\Theta_c$ of the trainable copy are trained with the conditioning input $c$. The trainable copy is connected to the LDM decoder block with two instances of zero convolutions, denoted $\mathcal{Z}(\cdot;\cdot)$ with parameters $\Theta_{\mathrm{z}1}$ and $\Theta_{\mathrm{z}2}$ respectively. Therefore, the finally ControlNet output $\boldsymbol{y}_{\boldsymbol{c}}$ is:

\begin{equation}\label{eqn-2}
    \boldsymbol{y}_{\boldsymbol{c}}=\mathcal{F} \left( \boldsymbol{x};\Theta \right) +\mathcal{Z} \left( \mathcal{F} \left( \boldsymbol{x}+\mathcal{Z} \left( \boldsymbol{c};\Theta _{\mathrm{z}1} \right) ;\Theta _{\mathrm{c}} \right) ;\Theta _{\mathrm{z}2} \right) 
\end{equation}

\subsection{Base Generation Stage}
\label{sec:1st_stage}





In the base generation stage, TexControl aims to generate the outline previews from input sketches and text. To capture detailed input sketch contour information, we employ the ControlNet scribble as our first-stage generative model, which is trained with human scribbles and can generate images following the input sketches faithfully. However, the sketch contains sparse and ambiguous information that cannot accurately constrain the model's inference process. To mitigate the issue of diverse generated results, we additionally utilize text prompts as constraint conditions during generation.

We simplify the Equation~\ref{eqn-2} in the following form: 
\begin{equation}\label{cn_sim}
    \boldsymbol{y}_{\boldsymbol{c}}=\mathcal{N} \left( \boldsymbol{x}, \boldsymbol{c} \right)
\end{equation}
where $\mathcal{N}$ is the ControlNet. Thus, the generation process in the base generation stage can be expressed specifically as:
\begin{equation}\label{cn_s2i}
    \boldsymbol{y}_{\boldsymbol{o}}=\mathcal{N}_{\boldsymbol{s\text{-}i}} \left( \boldsymbol{s}, \boldsymbol{p} \right)
\end{equation}
Where ${y}_{\boldsymbol{o}}$ is the middle outline previews, $\mathcal{N}_{\boldsymbol{s\text{-}i}}$ is the sketch-to-image ControlNet scribble model, $\boldsymbol{s}$ is the input sketches, $\boldsymbol{p}$ is the text prompts.

\subsection{Texture Control Stage}
\label{sec:2nd_stage}




In the texture control stage, our objective is to generate clothing images with detailed textures and specified materials from contour previews, thereby completing the entire fashion design process. In particular, the outline information contained in the outline previews needs to be strictly followed and introduced to the results. In addition, we also use text prompts in this stage to fine-grained control of the textures and materials.

We applied the ControlNet ip2p, ControlNet scribble, and LDM to accomplish this task, which can be expressed as:

\begin{equation}\label{f_i2i}
   \boldsymbol{y}_{\boldsymbol{r}}=\mathcal{N}_{\boldsymbol{i\text{-}i}} \left( \boldsymbol{y}_{\boldsymbol{o}}, \boldsymbol{p} \right)
\end{equation}
Where $\boldsymbol{y}_{\boldsymbol{r}}$ is the fine-designed results and $\mathcal{N} _{\boldsymbol{i\text{-}i}}$ is the image-to-image model based on ControlNet ip2p.

We use the model merge method to change the checkpoints of $\mathcal{N} _{\boldsymbol{i\text{-}i}}$, improving the generation results. Model merge is a model application method based on diffusion models, by adjusting the merging weights to integrate the pre-trained weights of U-Net in several pre-trained checkpoints, which can integrate the visual features of several pre-trained models. The final generated effect can have the visual effects of multiple models. In the model merge method, the fused input model has three weight parameters: input $\boldsymbol{w}_{\boldsymbol{i}}$, middle $\boldsymbol{w}_{\boldsymbol{m}}$, and output $\boldsymbol{w}_{\boldsymbol{o}}$. The $\boldsymbol{w}_{\boldsymbol{i}}$ affects the extraction of features in the downsampling process, and the $\boldsymbol{w}_{\boldsymbol{m}}$ produces more influence for the fused model under conditions similar to the $\boldsymbol{w}_{\boldsymbol{i}}$. The  $\boldsymbol{w}_{\boldsymbol{o}}$ affects the restoration of features during the upsampling process. The additional model influences the final model higher when the weight values are higher. The checkpoints of model $ \mathcal{N} _{\boldsymbol{i\text{-}i}}$ can expressed as:

\begin{equation}\label{n_i2i}
    \mathcal{C} _{\boldsymbol{i\text{-}i}}=\boldsymbol{w}*\mathcal{C} _{\boldsymbol{scr}}+\left( \boldsymbol{1}-\boldsymbol{w} \right)* \mathcal{C} _{\boldsymbol{ldm}}
\end{equation}


Where $\mathcal{C} _{\boldsymbol{scr}}$
is the ControlNet scribble checkpoints, $\mathcal{C} _{\boldsymbol{ldm}}$ is the LDM checkpoints, and $\boldsymbol{w}\in \left[ 0,1 \right] $ is the merge weight. It will be $\boldsymbol{w}_{\boldsymbol{i}}$, $\boldsymbol{w}_{\boldsymbol{m}}$, and $\boldsymbol{w}_{\boldsymbol{o}}$ in the corresponding network layers.

\section{Experiments and Results}

We conduct qualitative experiments to validate the image quality and sketch consistency of our model's results. In Section \ref{section:im} we introduce the implementation details of our experiment. We present the design and results of our qualitative evaluations in Section \ref{section:qe}. We also plan to implement quantitative experiments to verify our method as soon as possible.

\subsection{Implementation Details}
\label{section:im}




\begin{figure}[h]
    \centering
    \includegraphics[width=0.95\linewidth]{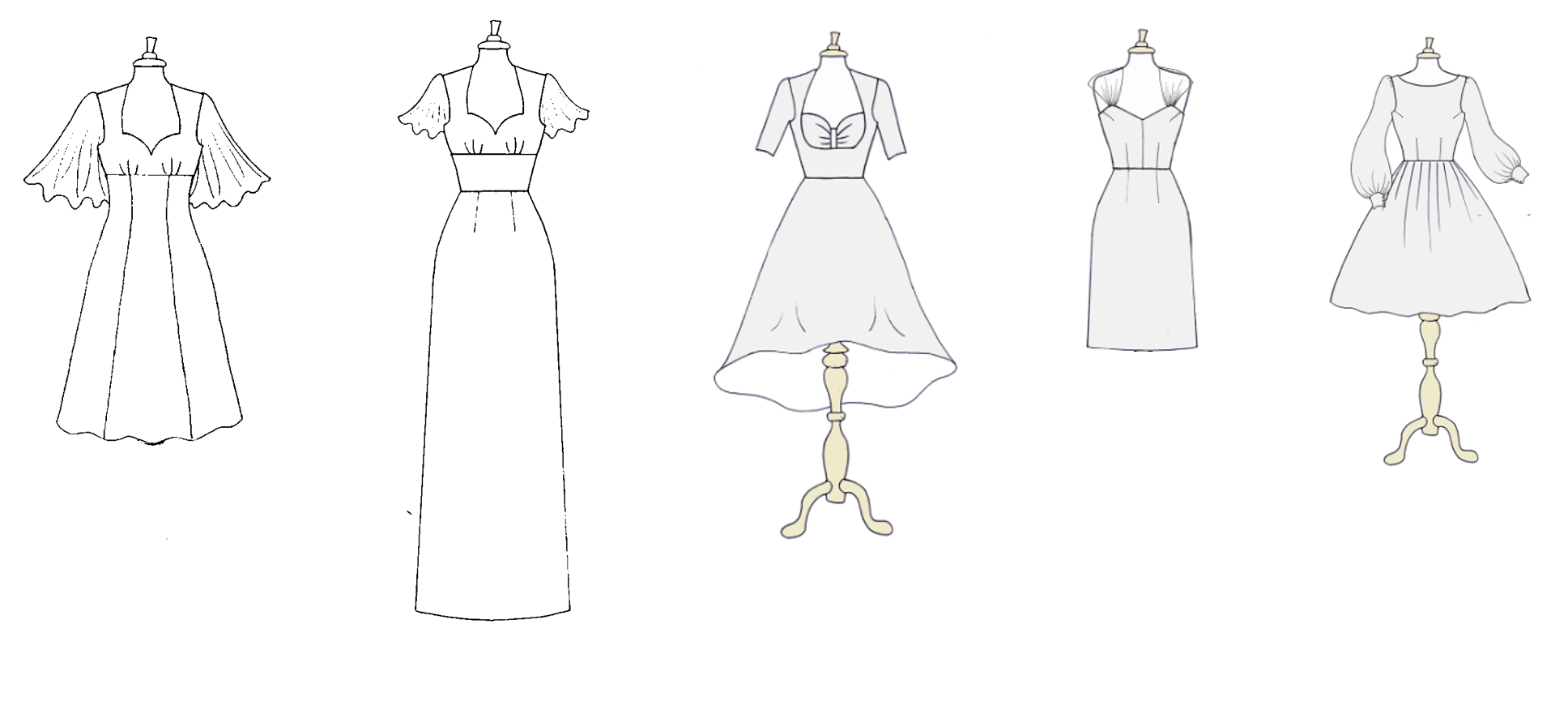}
    \caption{We collected diverse sketches through various sources and approaches.}
    \label{fig:sketch}
\end{figure}

We use the ControlNet scribble model (trained on the Synthesized scribbles dataset) in the base generation stage. And, we use the ControlNet ip2p model (trained on the Instruct Pix2pix dataset), ControlNet scribble model, and LDM in the texture control stage.
We use hand-drawn sketches of fashion show images, design sketches from the Dig for Victory Clothing website\footnote{https://www.digforvictoryclothing.com/design/your/own/dress} and other open-source images. Fig.~\ref{fig:sketch} shows part of our collection of sketches from the website. We collected a total of 50 different evening dress sketches as the main test subjects. 
We manually sharpened and high-definition the sketches to make them suitable for the generative models.


\begin{figure}[h]
    \centering
    \includegraphics[width=0.95\linewidth]{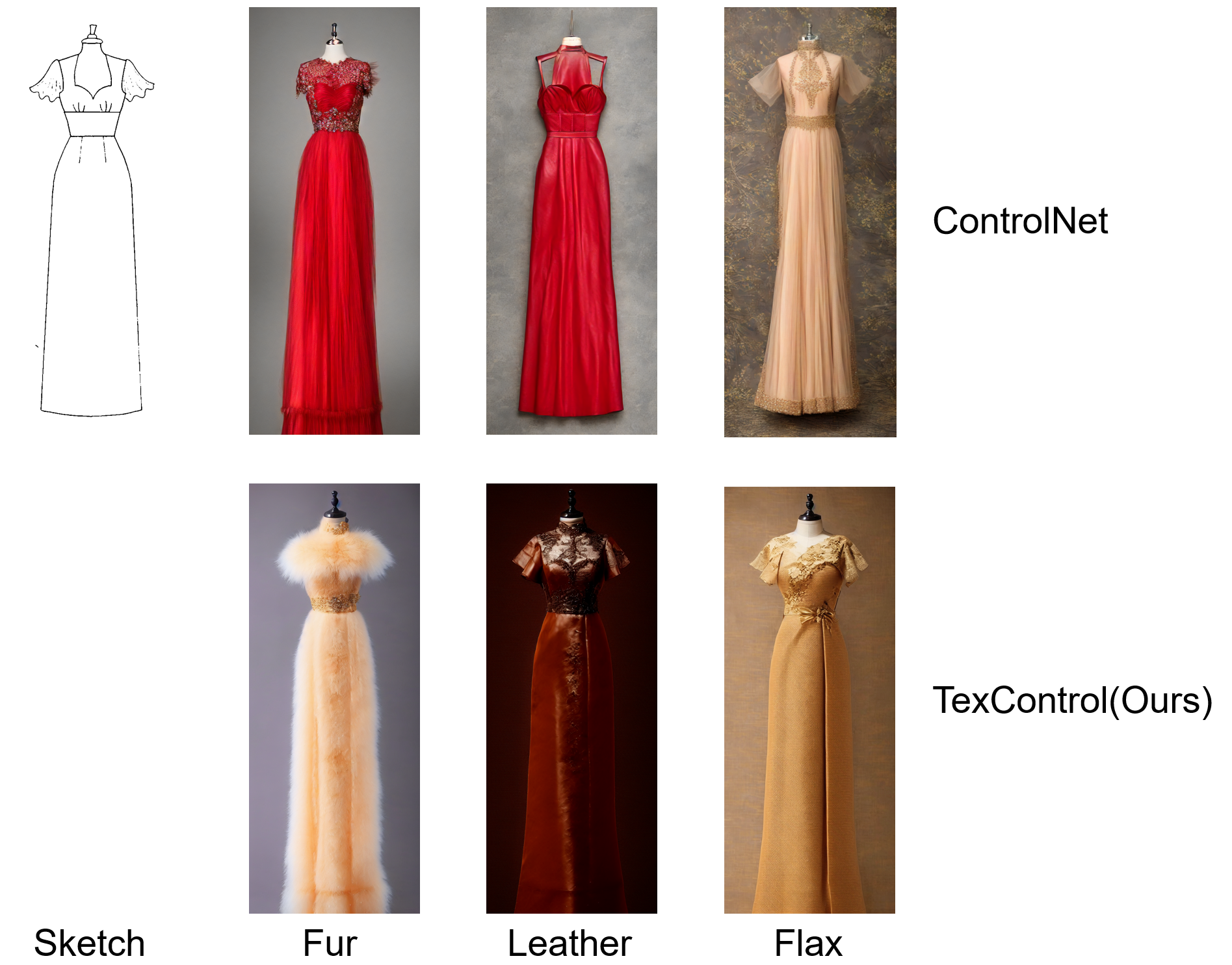}
    \caption{The result compare with the TexControl(Ours) and ControlNet.}
    \label{fig:result_1}
\end{figure}

The generation time step is set as $40$ in the base generation stage, and $80$ in the texture control stage. The prompt words, such as ``A photo of an evening dress", control the basic direction of the generation in the base generation stage. In the texture control stage, prompt words such as ``Only change the texture of the evening dress to Fur cloth" control the texture generation. The pre-trained checkpoint is the merge model. The main model we use in the model merge step is LDM-V1.5 and use the ControlNet scribble model again as the added model, the $\boldsymbol{w}_{\boldsymbol{i}}$ is $0.4$, the $\boldsymbol{w}_{\boldsymbol{m}}$ is $0.5$, and the $\boldsymbol{w}_{\boldsymbol{o}}$ is $1.0$. And the ControlNet part we use is ControlNet ip2p. 
Additionally, we made a distinction between synonyms, such as using ``fur cloth" instead of simply ``fur". In the end, we got over 500 generated images.

\subsection{Qualitative Evaluation}
\label{section:qe}




To illustrate the difference between our model and the state-of-the-art (SOTA) mainstream text-to-image model, we use the initial ControlNet scribble model\cite{zhang2023adding} as a reference.


As shown in Fig.~\ref{fig:result_1}, the SOTA sketch-to-image diffusion model obtains global information and generates an image from the text and sketch, but it cannot further control the texture and materials of the generated image. Our model undergoes two-stage control, ensuring not only faithful adherence to the input sketch outlines but also compliance with the specified texture and material information. 



\begin{figure}[h]
    \centering
    \includegraphics[width=0.95\linewidth]{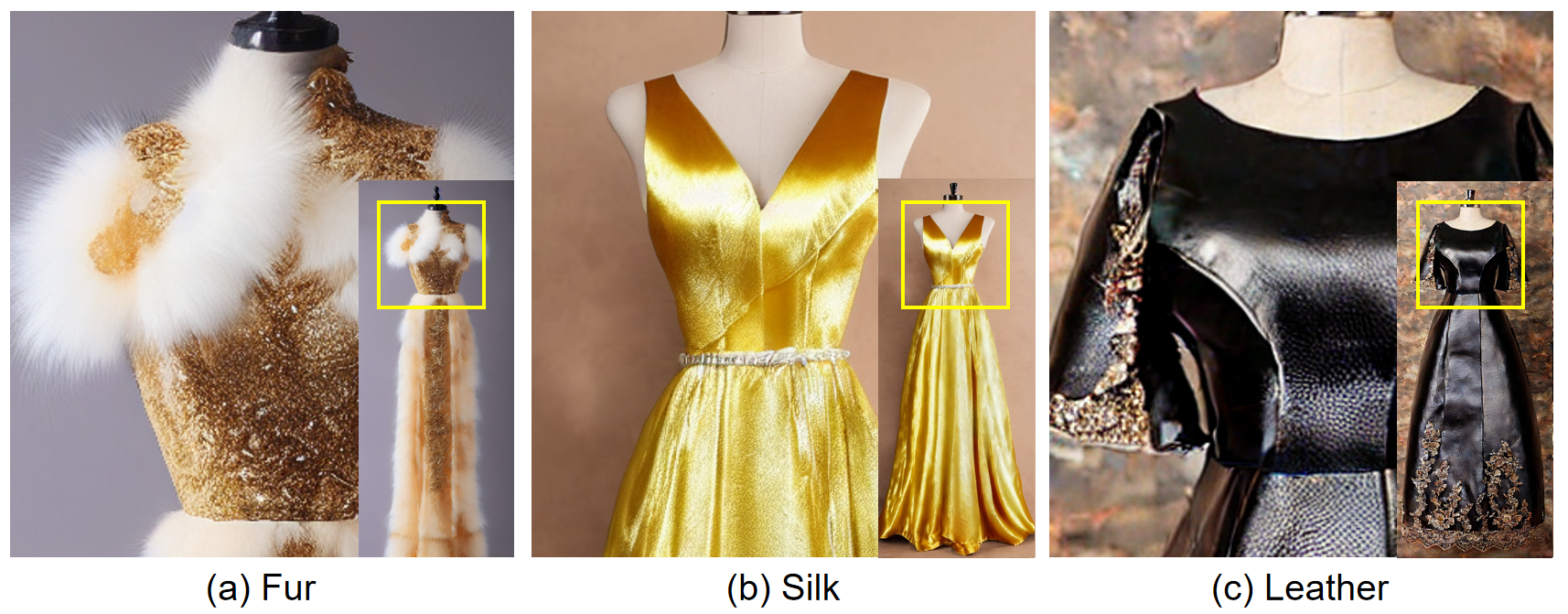}
    \caption{TexControl is good at generating fine-grained texture.}
    \label{fig:result_3}
\end{figure}

TexControl decomposes the complex control task into outline control and texture control using a two-stage model, thereby resulting in generated images that closely resemble real fashion design results. As shown in Fig.~\ref{fig:result_3}, TexControl is good at generating fine-grained texture with a reasonable outline shape guidance.

\begin{figure}[h]
    \centering
    \includegraphics[width=0.95\linewidth]{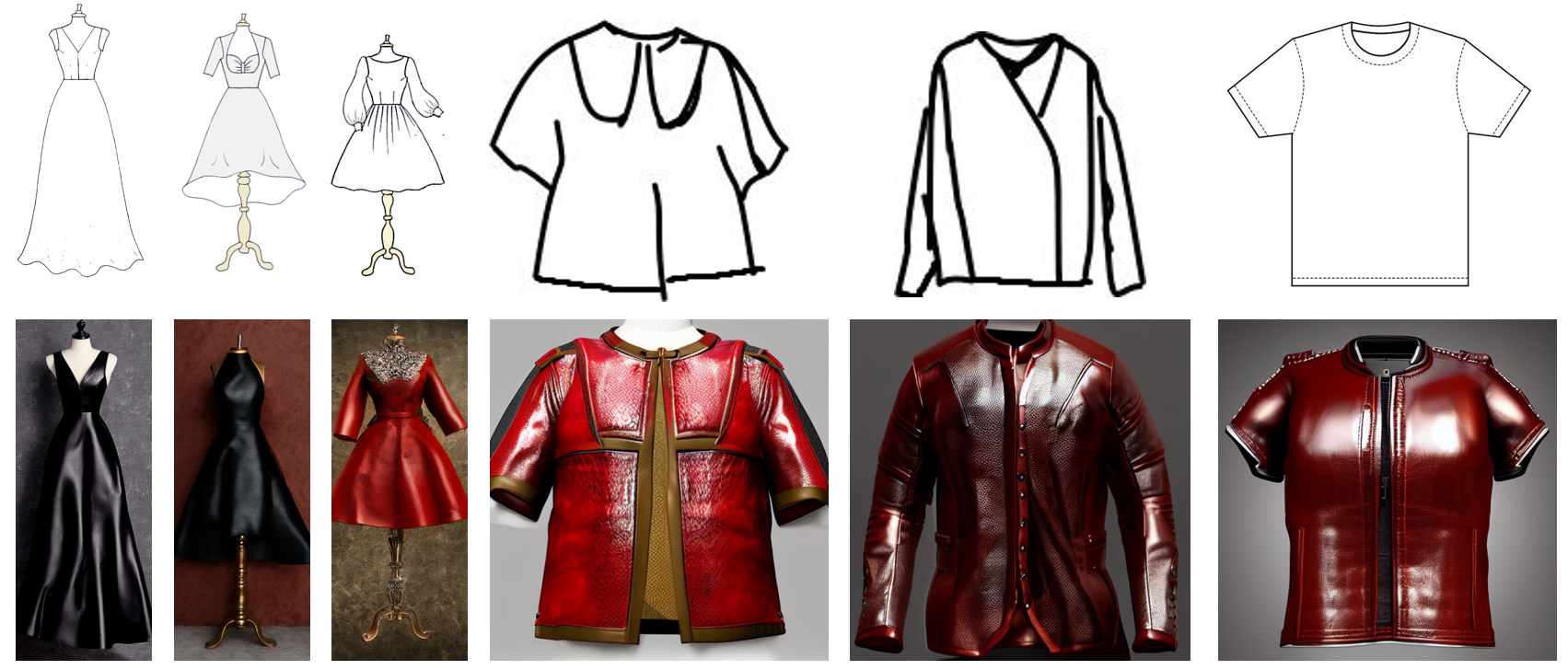}
    \caption{Leather clothing generation results from different clothing types and different styles of sketches.}
    \label{fig:result_4}
\end{figure}

\begin{figure}[h]
    \centering
    \includegraphics[width=0.95\linewidth]{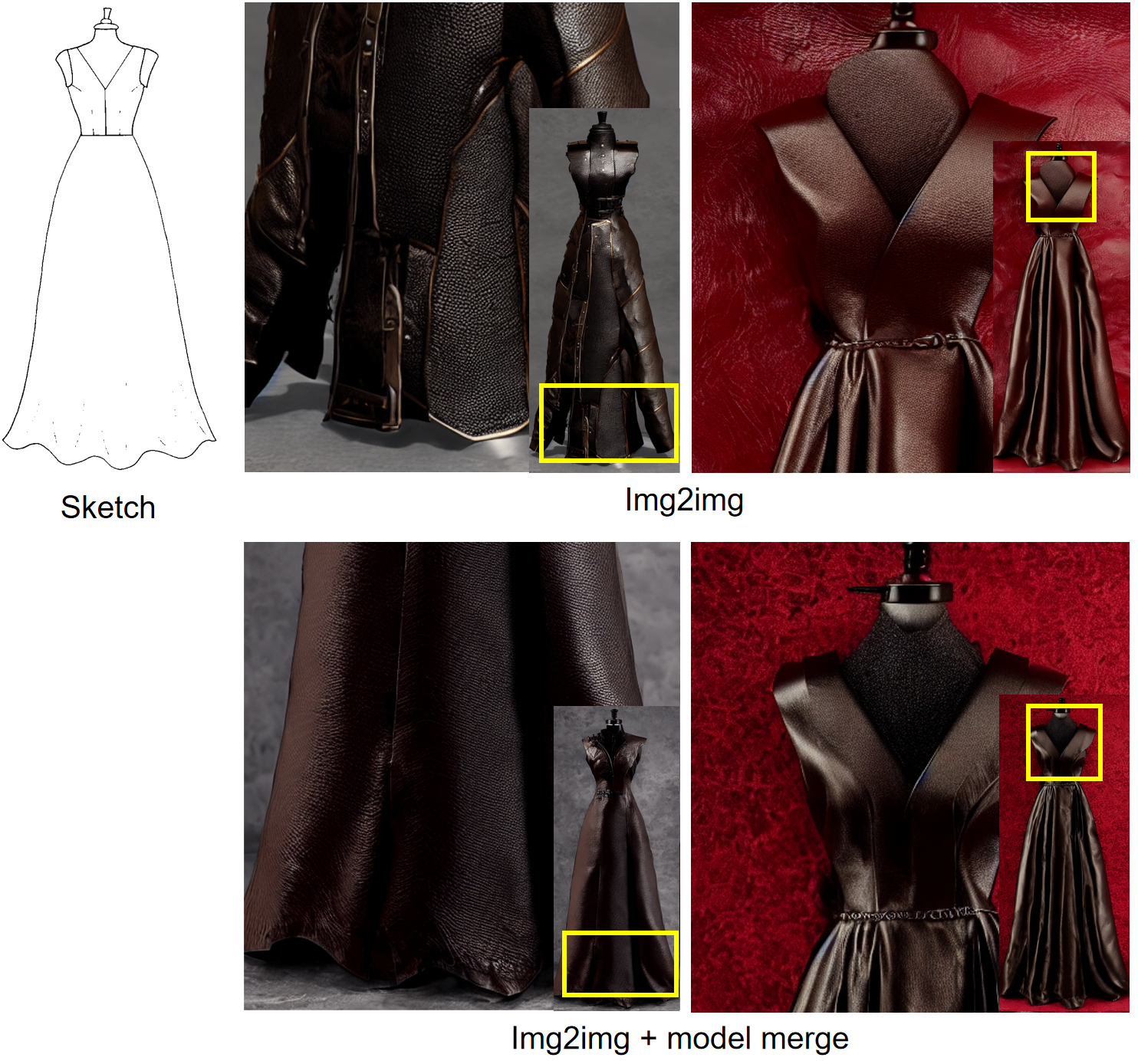}
    \caption{Comparison of model merge effects. The skirt hems and collars of generated results are more consistent with the input sketches.}
    \label{fig:result_7}
\end{figure}

We also present more results from diversity input sketches in Fig.~\ref{fig:result_4}, these results show the accuracy of texture. Even when the input sketches contain misleading information such as hangers, TexControl is still able to generate reasonable results. As shown in Fig.~\ref{fig:result_7}, in the texture control stage, the results with the model merge have better skirt hems and collars. 

\begin{figure}[h]
    \centering
    \includegraphics[width=0.95\linewidth]{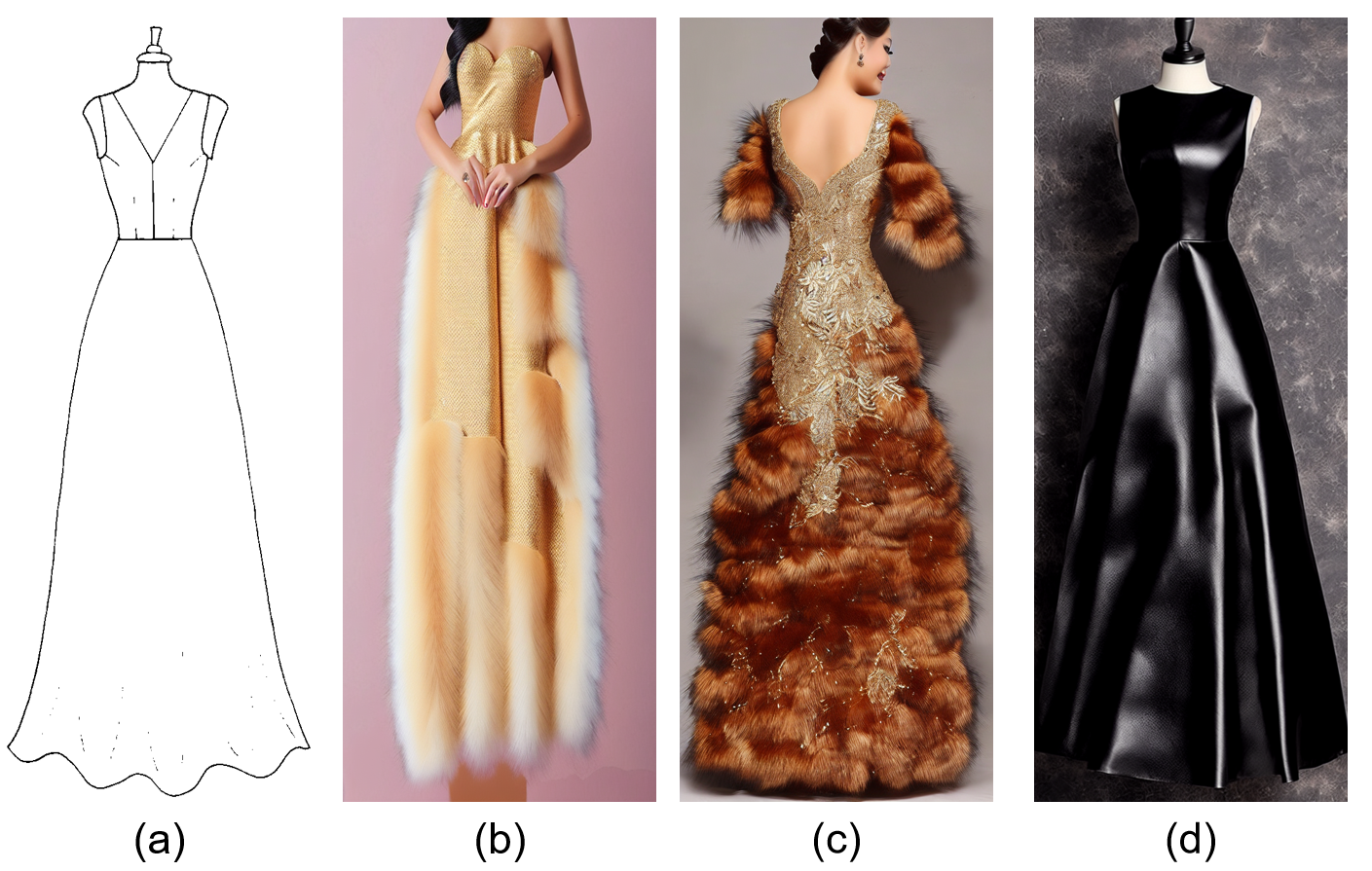}
    \caption{Failure results. When the input is sketch (a), the output (b) and (c) show that the algorithm generates the human body by accident. The collar in (d) does not match that required by sketch (a).}
    \label{fig:failure_result_01}
\end{figure}


\section{Conclusion}



This work proposed TexControl, a two-stage fashion image generation model based on ControlNet,  to help users design clothes intuitively and accurately. TexControl enables even novices to achieve precise texture and material designs. Finally, we conducted qualitative evaluations to verify our proposed method, that TexControl can generate more accurate texture and more complex materials than ControlNet.

TexControl still fails in some generation tasks. As shown in Fig.~\ref{fig:failure_result_01} (b) and (c), in the partial generation process, due to the strong correlation between the pre-training model's clothing and the human body, the human body will also be generated by accident. And, the collar in (d) does not match that required by sketch (a). These types of problems appear in the texture control stage, and these problems are also present in the ControlNet model.

For future work, we will conduct the quantitative evaluation in future experiments. We can use Frechet Inception Distance score (FID)\cite{heusel2017gans}, and may propose a new metric to evaluate the generative model from outline, texture, and detail. And, for the mistake generated, we can build a clothing-only dataset and use this dataset to train a new ControlNet model to weaken the relationship between the human body and clothing. 

\section*{Acknowledgment}
This research has been supported by JSPS KAKENHI Grant Number 23K18514, and the Kayamori Foundation of Informational Science  Advancement. We thank the anonymous reviewers for their insightful comments.


\bibliographystyle{ieeetran}
\bibliography{ref}

\begin{thebibliography}{10}
\providecommand{\url}[1]{#1}
\csname url@samestyle\endcsname
\providecommand{\newblock}{\relax}
\providecommand{\bibinfo}[2]{#2}
\providecommand{\BIBentrySTDinterwordspacing}{\spaceskip=0pt\relax}
\providecommand{\BIBentryALTinterwordstretchfactor}{4}
\providecommand{\BIBentryALTinterwordspacing}{\spaceskip=\fontdimen2\font plus
\BIBentryALTinterwordstretchfactor\fontdimen3\font minus \fontdimen4\font\relax}
\providecommand{\BIBforeignlanguage}[2]{{%
\expandafter\ifx\csname l@#1\endcsname\relax
\typeout{** WARNING: IEEEtran.bst: No hyphenation pattern has been}%
\typeout{** loaded for the language `#1'. Using the pattern for}%
\typeout{** the default language instead.}%
\else
\language=\csname l@#1\endcsname
\fi
#2}}
\providecommand{\BIBdecl}{\relax}
\BIBdecl

\bibitem{he2023sketch2cloth}
Y.~He, H.~Xie, and K.~Miyata, ``Sketch2cloth: Sketch-based 3d garment generation with unsigned distance fields,'' \emph{arXiv preprint arXiv:2303.00167}, 2023.

\bibitem{jain2019text}
A.~Jain, D.~Modi, R.~Jikadra, and S.~Chachra, ``Text to image generation of fashion clothing,'' in \emph{2019 6th International Conference on Computing for Sustainable Global Development (INDIACom)}.\hskip 1em plus 0.5em minus 0.4em\relax IEEE, 2019, pp. 355--358.

\bibitem{saharia2022photorealistic}
C.~Saharia, W.~Chan, S.~Saxena, L.~Li, J.~Whang, E.~L. Denton, K.~Ghasemipour, R.~Gontijo~Lopes, B.~Karagol~Ayan, T.~Salimans \emph{et~al.}, ``Photorealistic text-to-image diffusion models with deep language understanding,'' \emph{Advances in Neural Information Processing Systems}, vol.~35, pp. 36\,479--36\,494, 2022.

\bibitem{ramesh2022hierarchical}
A.~Ramesh, P.~Dhariwal, A.~Nichol, C.~Chu, and M.~Chen, ``Hierarchical text-conditional image generation with clip latents,'' \emph{arXiv preprint arXiv:2204.06125}, 2022.

\bibitem{rombach2022high}
R.~Rombach, A.~Blattmann, D.~Lorenz, P.~Esser, and B.~Ommer, ``High-resolution image synthesis with latent diffusion models,'' in \emph{Proceedings of the IEEE/CVF Conference on Computer Vision and Pattern Recognition}, 2022, pp. 10\,684--10\,695.

\bibitem{zhang2023adding}
L.~Zhang, A.~Rao, and M.~Agrawala, ``Adding conditional control to text-to-image diffusion models,'' in \emph{Proceedings of the IEEE/CVF International Conference on Computer Vision}, 2023, pp. 3836--3847.

\bibitem{sohl2015deep}
J.~Sohl-Dickstein, E.~Weiss, N.~Maheswaranathan, and S.~Ganguli, ``Deep unsupervised learning using nonequilibrium thermodynamics,'' in \emph{International Conference on Machine Learning}.\hskip 1em plus 0.5em minus 0.4em\relax PMLR, 2015, pp. 2256--2265.

\bibitem{song2019generative}
Y.~Song and S.~Ermon, ``Generative modeling by estimating gradients of the data distribution,'' \emph{Advances in neural information processing systems}, vol.~32, 2019.

\bibitem{ho2020denoising}
J.~Ho, A.~Jain, and P.~Abbeel, ``Denoising diffusion probabilistic models,'' \emph{Advances in Neural Information Processing Systems}, vol.~33, pp. 6840--6851, 2020.

\bibitem{luo2023latent}
S.~Luo, Y.~Tan, L.~Huang, J.~Li, and H.~Zhao, ``Latent consistency models: Synthesizing high-resolution images with few-step inference,'' \emph{arXiv preprint arXiv:2310.04378}, 2023.

\bibitem{cui2018fashiongan}
Y.~R. Cui, Q.~Liu, C.~Y. Gao, and Z.~Su, ``Fashiongan: display your fashion design using conditional generative adversarial nets,'' in \emph{Computer Graphics Forum}, vol.~37, no.~7.\hskip 1em plus 0.5em minus 0.4em\relax Wiley Online Library, 2018, pp. 109--119.

\bibitem{jiang2022text2human}
Y.~Jiang, S.~Yang, H.~Qiu, W.~Wu, C.~C. Loy, and Z.~Liu, ``Text2human: Text-driven controllable human image generation,'' \emph{ACM Transactions on Graphics (TOG)}, vol.~41, no.~4, pp. 1--11, 2022.

\bibitem{baldrati2023multimodal}
A.~Baldrati, D.~Morelli, G.~Cartella, M.~Cornia, M.~Bertini, and R.~Cucchiara, ``Multimodal garment designer: Human-centric latent diffusion models for fashion image editing,'' \emph{arXiv preprint arXiv:2304.02051}, 2023.

\bibitem{heusel2017gans}
M.~Heusel, H.~Ramsauer, T.~Unterthiner, B.~Nessler, and S.~Hochreiter, ``Gans trained by a two time-scale update rule converge to a local nash equilibrium,'' \emph{Advances in neural information processing systems}, vol.~30, 2017.

\end{thebibliography}

\end{document}